\pdfoutput=1

\documentclass[11pt]{article}

\usepackage[final]{acl}

\usepackage{times}
\usepackage{latexsym}
\usepackage{amsmath}

\usepackage[T1]{fontenc}

\usepackage[utf8]{inputenc}

\usepackage{microtype}

\usepackage{inconsolata}

\usepackage{graphicx}
\usepackage{float}
\usepackage{listings}
\usepackage{xcolor}
\usepackage{colortbl}
\usepackage[most]{tcolorbox}
\usepackage{caption}

\title{Enhancing Interpretable Image Classification Through LLM Agents and Conditional Concept Bottleneck Models}

\author{
\textbf{Yiwen Jiang\textsuperscript{1,2}},
\textbf{Deval Mehta\textsuperscript{1,2}},
\textbf{Wei Feng\textsuperscript{1,2}},
\textbf{Zongyuan Ge\textsuperscript{2}}
\\
\textsuperscript{1}Faculty of Engineering, Monash University, Melbourne, Australia \\
\textsuperscript{2}AIM for Health Lab, Faculty of IT, Monash University, Melbourne, Australia \\
\texttt{\{yiwen.jiang, deval.mehta, wei.feng, zongyuan.ge\}@monash.edu}
}

\lstdefinestyle{plain}{
    basicstyle=\fontsize{7}{9.5}\ttfamily,
    keywordstyle=\color{blue},
    commentstyle=\color{gray},
    stringstyle=\color{green},
    showstringspaces=false,
    breaklines=true,
    breakatwhitespace=false,
    breakindent=0pt,
    escapeinside={(*@}{@*)}
}

\begin{document}
\maketitle
\begin{abstract}
Concept Bottleneck Models (CBMs) decompose image classification into a process governed by interpretable, human-readable concepts. Recent advances in CBMs have used Large Language Models (LLMs) to generate candidate concepts. However, a critical question remains: What is the optimal number of concepts to use? Current concept banks suffer from redundancy or insufficient coverage. To address this issue, we introduce a dynamic, agent-based approach that adjusts the concept bank in response to environmental feedback, optimizing the number of concepts for sufficiency yet concise coverage. Moreover, we propose Conditional Concept Bottleneck Models (CoCoBMs) to overcome the limitations in traditional CBMs' concept scoring mechanisms. It enhances the accuracy of assessing each concept’s contribution to classification tasks and feature an editable matrix that allows LLMs to correct concept scores that conflict with their internal knowledge. Our evaluations across 6 datasets show that our method not only improves classification accuracy by 6\% but also enhances interpretability assessments by 30\%.
\end{abstract}

\section{Introduction}

Deep Learning (DL) models have excelled in various fields, but their black-box nature limits the interpretability of their decision-making processes \citep{papernot2017practical}. Increasing attention has been directed toward developing intrinsically interpretable and flexible DL models. This research predominantly revolves around concept analysis \citep{kim2018interpretability}, which aims to understand how neural networks encode and utilize high-level, human-interpretable features. Concept Bottleneck Models (CBMs) \citep{koh2020concept} are among the most representative approaches in this direction, mapping visual representations to a set of human-understandable textual concepts, from which the final decision is derived through a linear combination of these concept scores.

Recent research on CBMs has established a new language grounding paradigm (\citealp{oikarinen2023label}; \citealp{yang2023language}; \citealp{yan2023learning}). It first prompts pre-trained Large Language Models (LLMs) with class names to generate candidate concept sets. Various concept selection algorithms are then designed to identify the most representative or distinguishing concepts. Finally, multimodal pre-trained models such as CLIP \citep{radford2021learning}, align visual features with textual descriptions by projecting visual representations into each concept embedding, forming a concept bottleneck layer.

This CLIP-based paradigm eliminates the need for manually constructing a concept bank and annotating each concept within the images. Concurrently, it retains the key advantage of CBMs by enabling human intervention, allowing users to directly edit erroneous concept scores to correct model behavior \citep{koh2020concept}. Although CBMs have improved the interpretability of image classification tasks, several unresolved challenges remain in grounding abstract concepts from LLMs to diverse and unpredictable downstream images.

First, CBMs have an inherent intrepretability and accuracy trade-off, but some CLIP-based CBMs have provided a comparable performance to the standard neural networks depending on the dataset. Another key challenge lies in determining the optimal number of concepts required for a concept bank. Previous studies typically rely on manually specifying the number of concepts and subsequently employing concept selection algorithms to construct a fixed-size concept bank. For instance, LaBo \citep{yang2023language} assigns $k$ concepts per category, resulting in a concept bank with 10,000 concepts for the CUB dataset \citep{wah2011caltech}, which includes 200 bird species. In contrast, LM4CV \citep{yan2023learning} adopts a significantly smaller concept bank with only 32 concepts, yet achieves competitive classification performance on the same dataset. While accuracy generally improves as the number of specified concepts increases, the ideal number of concepts remains an open question. Third, prior work (\citealp{oikarinen2023label}; \citealp{yan2023towards}) has demonstrated that humans can interact with CBMs by manually editing concept scores in the bottleneck layer to correct mispredictions and alter model behavior. These mispredictions often stem from concept activations that contradict objective facts. However, such edits have been primarily limited to the test-time setting \citep{koh2020concept, hu2024editable}. To date, no research has explored the use of LLMs' inherent factual knowledge to automatically edit incorrectly activated concepts in CBMs during training.

In this work, we propose a novel framework to holistically address the challenges introduced above. Our analysis reveals that the performance bottleneck of traditional CBMs primarily stems from a unified scoring mechanism across all categories. To address this issue, we introduce Conditional Concept Bottleneck Models (CoCoBMs) that incorporates category-specific scoring and weighting mechanisms to project visual information into category-conditioned concept embeddings. This forms a conditional concept bottleneck layer, significantly enhancing the model's performance.

Moreover, existing studies follow a static language grounding paradigm \citep{chandu2021grounding}, which makes it difficult to determine the optimal number of concepts. In a one-directional workflow, LLMs generate concepts for CBMs without interaction with downstream visual data, thereby missing valuable feedback for refining grounded concepts. To incorporate feedback, our proposed framework incorporates a Concept Agent that leverages few-shot feedback to analyze concept activation patterns from downstream image data, enabling the identification of redundant and insufficient concepts. By dynamically refining and expanding the concept bank, the Agent automates the determination of the optimal concept count. Furthermore, the Concept Agent is endowed with global editing authority over CoCoBMs' activation scores, enabling it to identify and suppress activations that conflict with the factual knowledge encoded within LLMs.

Furthermore, we develop a quantitative metric to evaluate the interpretability of model predictions by converting conceptual evidence into textual descriptions. These descriptions are then assessed by LLMs in terms of truthfulness and distinguishability. Evaluations across 6 datasets validate the effectiveness of our approach in terms of both classification accuracy and interpretability. Overall, our main contributions are as follows: 

$\bullet$ We propose Conditional Concept Bottleneck Models (CoCoBMs), which incorporate category-specific scoring and weighting mechanisms, to enhance the model's classification performance.

$\bullet$ We propose a Concept Agent that dynamically grounds the concept bank by using environmental feedback to identify and refine redundancies and gaps, optimizing the concept count.

$\bullet$ We conduct evaluations on 6 datasets, demonstrating a 6\% increase in classification accuracy, and around a 30\% improvement in interpretability through our designed quantitative assessment.

\section{Related Work}

\noindent \textbf{Concept Bottleneck Models.} CBMs \citep{koh2020concept} are a prominent approach for designing inherently interpretable DL models, as detailed by \citet{zhou2018interpretable} and \citet{losch2019interpretability}. CBMs incorporate a concept bottleneck layer preceding the final fully connected layer, where each neuron represents a human-interpretable concept. Some variants of CBMs have been developed to mitigate inherent drawbacks. For example, \citet{yuksekgonul2023posthoc} and \citet{oikarinen2023label} proposed data-efficient methods to convert any DL models into CBMs without training from scratch. By leveraging multimodal pre-trained models \citep{fong2018net2vec} to learn concept activations, they bypassed the necessity for concept annotations. CBMs enable model debugging and analysis by allowing edits to concept scores or weights, optimizing single-sample predictions or global behavior (\citealp{koh2020concept}; \citealp{oikarinen2023label}; \citealp{yan2023learning}). However, this process often demands significant human effort, limiting its scalability.

\noindent \textbf{Concept Bank Construction.} Recent efforts such as Label-free CBMs \citep{oikarinen2023label}, LaBo \citep{yang2023language} and LM4CV \citep{yan2023learning} have resorted to generate concepts by tapping into the knowledge base of LLMs \citep{brown2020language}. LaBo selected a fixed number of concepts for each category, while LM4CV proposed a learning-to-search approach to construct a concise bank covering all categories. On CUB dataset \citep{wah2011caltech}, they built concept banks with scales differing by several orders of magnitude, highlighting the question of what constitutes an optimal number of concepts for a bank. LaBo overlooks shared concepts across labels, inevitably resulting in redundancy. In contrast, LM4CV emphasizes conciseness but suffers from insufficiency. Recent studies (\citealp{yan2023learning}; \citealp{shang2024incremental}) indicate that a sufficiently large bank, even when constructed from randomly selected words can achieve accuracy comparable to that of an interpretable one. A reasonable number of concepts should lie between $\log_2 n$ and $d$, where $n$ is the number of categories and $d$ is the dimensionality of the concept embeddings. Furthermore, some work, such as P-CBM \citep{yuksekgonul2023posthoc} and Res-CBM \citep{shang2024incremental}, retrieves concepts from Knowledge Graphs (KG) \citep{speer2017conceptnet}, which heavily depend on how the KG are built. While Res-CBM trys to address insufficiency, it remains limited to static KG, complementing selection algorithms. No work has adopted a dynamic grounding paradigm for concept bank construction and refinement.

\noindent \textbf{LLM-based Autonomous Agents.} Autonomous agents aim to achieve AGI through self-directed planning and actions \citep{wang2024survey}. Recent advances in Chain-of-Thought (CoT) reasoning \citep{wei2022chain} have positioned LLMs as central controllers, enabling human-like decision-making by integrating perception, memory, and action capabilities. LLM-based agents typically follow a unified framework comprising three modules: memory, planning, and action (\citealp{yao2022react}; \citealp{zhu2023ghost}; \citealp{huang2022inner}). The memory module stores information to aid future planning, while the planning module deconstructs tasks, often using feedback from environmental interactions to enable self-evolution. The action module executes decisions, directly interacting with and impacting the environment. The research community has not explored employing such LLM-based agents for concept-based interpretable image classification.

\begin{figure}[t]
  \includegraphics[width=\columnwidth]{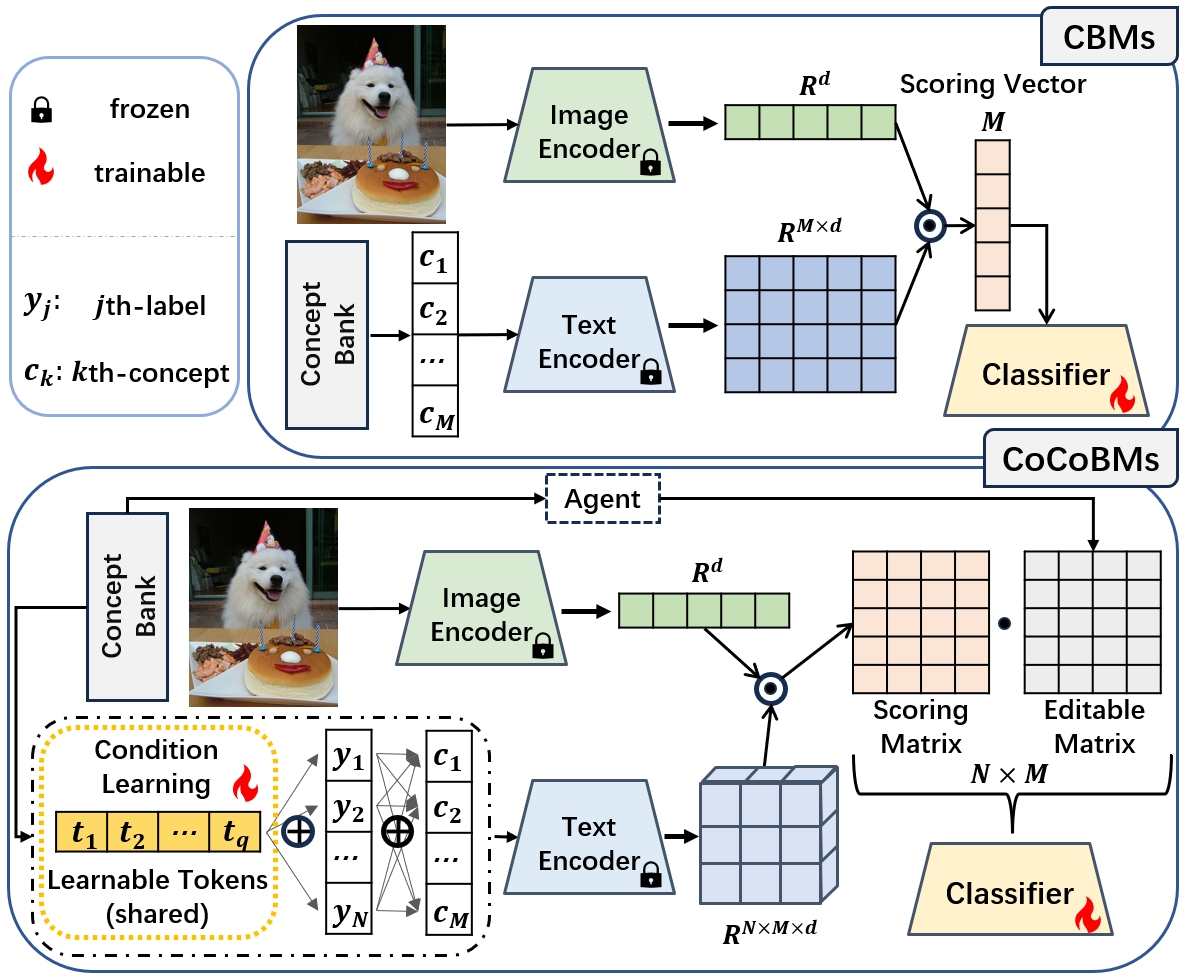}
  \caption{Architectural comparison of CBMs and CoCoBMs. CoCoBMs employ label-conditioned scoring to enable category-specific concept evaluation. An editable matrix is introduced during training, allowing the agent to suppress incorrectly activated concepts.}
  \label{fig:cocobms}
\end{figure}

\section{Methodology}

\subsection{Conditional Concept Bottleneck Models}

\noindent \textbf{Problem Formulation.} Consider a dataset of image-label pairs $\mathcal{D}=\{(x_{i},y_{i})\}$, where each image $x_i \in \mathcal{X}$ is associated with a label $y_i \in \mathcal{Y}$ drawn from $N$ predefined categories. To facilitate interpretable classification, a set of $M$ semantic concepts $\mathcal{C}=\{c_{1},c_{2},\dots,c_{M}\}$ is introduced as an intermediate representation. Original CBMs \citep{koh2020concept} decompose prediction as $\hat{y}=f(g(x))$, where $g:\mathcal{R}^d \to \mathcal{R}^M$ maps image features to concept scores, and $f:\mathcal{R}^M \to \mathcal{Y}$ aggregates these scores into a label prediction.

Recent CBMs build on Visual-Language Models (VLMs), such as CLIP \citep{radford2021learning}, which consist of an image encoder $\mathcal{I}:\mathcal{X} \to \mathcal{R}^d$ and a text encoder $\mathcal{T}:\mathcal{C} \to \mathcal{R}^d$, projecting images and text into a shared $d$-dimensional feature space. Given an image $x_i$ and a concept set $\mathcal{C}$, CLIP-based CBMs \citep{yuksekgonul2023posthoc} compute concept scores $\vec{s_c} = [s_{c_1}, s_{c_2}, \dots, s_{c_M}]$, where each score is given by the dot product $s_{c_k} = \mathcal{I}(x_i) \cdot \mathcal{T}(c_k)$, measuring the cross-modal alignment. These concept scores are then aggregated into label-level scores $S_{\mathcal{Y}} = [s_y^1, s_y^2, \dots, s_y^N]$ via a learned concept weight matrix $W \in \mathcal{R}^{N \times M}$ that captures the relative importance of each concept for each label. This process adheres to a shared scoring mechanism, where the same set of concept scores $\vec{s_c}$ is reused across all labels $\mathcal{Y}$:
\begin{equation}
  \label{eq:ori_cbm_score}
\vec{s_c} = P(\vec{s_c} \mid x_i, \mathcal{C}), \quad S_{\mathcal{Y}} = \big\Vert_{j=1}^N P(s_y^j \mid \vec{s_c})
\end{equation}
where $\big\Vert$ denotes the concatenation of per-label scores into the final prediction vector $S_{\mathcal{Y}}$. However, this formulation assumes an overly equitable sharing of concept scores across labels, overlooking the fact that a single concept may contribute unevenly to different categories.

\begin{figure*}[t]
  \centering
  \includegraphics[width=0.9\textwidth]{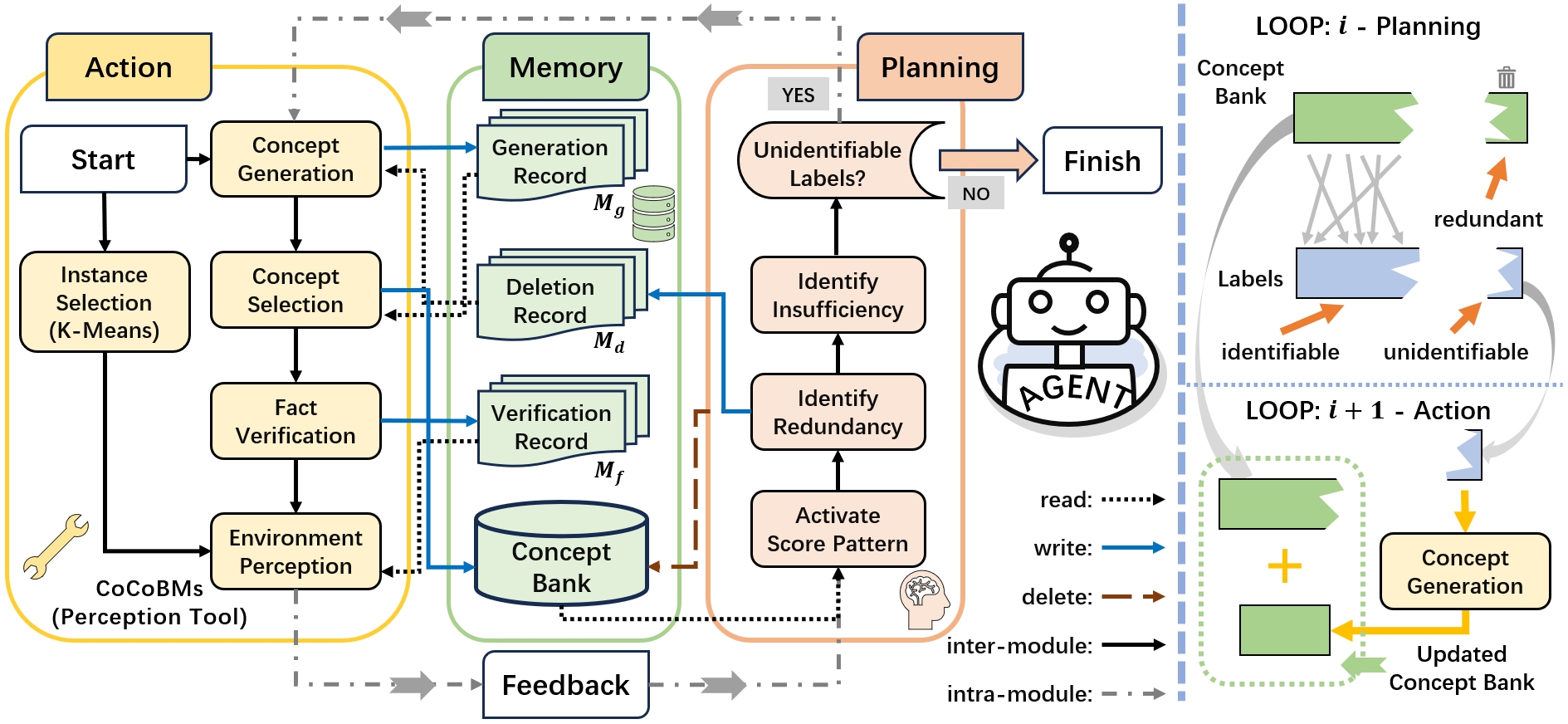} \hfill
  \caption {\textit{\textbf{Left}}: Modular components with intra-module and inter-module workflows in the Concept Agent. \textit{\textbf{Right}}: The planning module informs the action module to iteratively generate and refine concepts based on feedback.}
  \label{fig:main_figure}
\end{figure*}

\noindent \textbf{Category-Specific Scoring.} To address this limitation, we propose a category-specific scoring mechanism in CoCoBMs, as illustrated in Figure~\ref{fig:cocobms}, that replaces the shared concept scores with label-specific ones. The computation of $S_{\mathcal{Y}}$ is redefined as:
\begin{equation}
\begin{aligned}
  \vec{s_c^j} &= \big\Vert_{k=1}^M P(s_{c_k}^j \mid x_i, y_j, c_k) \\
  S_{\mathcal{Y}} &= \big\Vert_{j=1}^N P(s_y^j \mid \vec{s_c^j})
\end{aligned}
\label{eq:cocobm_score}
\end{equation}
where $s_{c_k}^j$ denotes the score of concept $c_k$ conditioned on the image $x_i$ and the hypothesized label $y_j$, and $\vec{s_c^j}$ is the resulting label-specific concept score vector. In contrast to original CBMs, which use a shared concept bottleneck across all labels, our formulation yields a label-specific concept matrix. Collapsing this matrix along the label dimension recovers the original CBM formulation, making it a special case of our method.

\noindent \textbf{Condition Learning.} CoCoBMs incorporate labels as conditional inputs during the concept scoring process. To achieve this, we adopt a prompt-learning strategy \citep{zhou2022learning, mehta2025interpretable}, in which learnable condition prompts are appended to the textual input, as illustrated in Figure~\ref{fig:cocobms}. Specifically, for the concept score $s_{c_k}^j$, the input text $p_k^j$ is constructed as:
\begin{equation}
  \label{eq:prompt_learning}
  p_k^j=\left[t_1\right]\left[t_2\right] \dots \left[t_q\right]\left[y_j\right]\left[c_k\right]
\end{equation}
where $\left[y_j\right]$ and $\left[c_k\right]$ denote the tokenized category and concept name, respectively, while each $t_i \in \{\ 1, 2, ... q \}$ is a learnable vector with the same dimensionality as CLIP word embeddings. These learnable tokens are shared across all labels and concepts to prevent information leakage. The final score is computed as $\mathcal{I}(x_i) \cdot \mathcal{T}(p_k^j)$, which constitutes an entry in the overall concept matrix $R^{N \times M}$.

\noindent \textbf{Editable Matrix.} CBMs provide interactivity \citep{koh2020concept} through editable scores and weights, but may activate concepts that contradict factual knowledge \citep{oikarinen2023label}. We propose an editable matrix $E$ to constrain false positive concepts $c_k$ associated with label $y_j$, defined as:
\begin{equation}
  \label{eq:editable_matrix}
    E_{jk} =
  \begin{cases} 
    1, & \text{if } c_k \notin y_j, \\ 
    0, & \text{if } c_k \in y_j.
  \end{cases}
\end{equation}
where $c_k \notin y_j$ indicates that concept $c_k$ is factually incompatible with category $y_j$ under any circumstances. The matrix $E$ encodes the factual relevance of each concept-label pair, determined automatically by our LLM-based Concept Agent (described in Section~\ref{sec:action-module}). To suppress factual false positives, we enforce:
\begin{equation}
    \label{eq:editable_score}
    s_{c_k}^j = \min(s_{c_k}^j, 0), \quad \text{where } E_{jk} = 1
\end{equation}
which sets the concept score to zero for any label-concept pair deemed invalid by the editable matrix.

\noindent \textbf{Objective Function.} The model is trained using a binary cross-entropy loss computed for each sample:
\begin{equation}
    -\frac{1}{N} \sum_{j=1}^N 
    \Big[W_p y_i \log(\hat{y}_i)+(1 - y_i) \log(1-\hat{y}_i)
    \Big]
\end{equation}
where $N$ is the number of labels, $y_i$ is the ground-truth label for input image $x_i$, and $\hat{y}_i = \sigma(s_y^j)$ is the predicted probability after applying the Sigmoid activation. The positive class weight $W_p = N$ compensates for label imbalance within each sample.

\subsection{Overview of the Concept Agent}
The Concept Agent is designed to construct a concept bank tailored to downstream image data, aiming to ensure sufficiency while minimizing redundancy by automatically optimizing the number of concepts. Figure~\ref{fig:main_figure} depicts the structure and overall workflow of the proposed agent, which comprises three key modules: \textit{memory}, \textit{action}, and \textit{planning}.

The \textit{action module} equips the agent with visual perception, enabling direct interaction with the environment. It is responsible for generating, selecting, and verifying concepts, as well as choosing instances that serve as feedback environments via CoCoBMs. The \textit{planning module} processes feedback to evaluate concept-label associations, removes redundant concepts, and supplements missing ones by guiding the action module. The \textit{memory module} logs interaction history and maintains versioned updates of the concept bank. The agent refines the concept bank iteratively until all labels can be reliably identified after eliminating redundancies.

\subsection{Memory Module}

The memory module maintains structured lists of generated concepts ($M_g$), deleted concepts ($M_d$), and fact-verified concept-label pairs ($M_f$). It also stores the updated concept bank after each iteration. This design enables the agent to perform read, write, and delete operations during action execution and planning, providing long-term, traceable memory to support iterative refinement.

\subsection{Action Module}
\label{sec:action-module}

\noindent \textbf{Concept Generation.} We prompt LLMs with category names to generate candidate concept lists. The prompt template is as follows (omitting detailed instructions on concept constraints and output format):  \textit{What are the helpful visual features to distinguish \texttt{[CLS]} from other \texttt{[S-CLS]}?} Here, \texttt{[CLS]} denotes the category name and \texttt{[S-CLS]} refers to its superclass (if known), or general object categories otherwise. For example, in the CUB dataset, the class \texttt{[CLS]} may be \textit{Cardinal}, while the superclass \texttt{[S-CLS]} is \textit{bird}. To avoid duplicate concepts during iteration, if a deleted concept $c_j \in M_d$ was previously generated by the same \texttt{[CLS]} prompt, it is appended to the prompt to prevent regeneration.

\noindent \textbf{Concept Selection.} This action selects a fixed number of concepts from the candidate pool to augment the current concept bank. We adopt the learning-to-search method proposed by \citet{yan2023learning}, which learns a dictionary to approximate a subset of concepts \citep{van2017neural}, and applies a classification head to project the dictionary onto $N$ labels, trained by categorical cross-entropy loss. In our setting, if the planning module identifies a subset $n \in N$ of unidentifiable labels, the classification head is modified to predict $|n|+1$ classes, where $N \setminus n$ is grouped into a single negative class.

\noindent \textbf{Fact Verification.} It verifies each concept-label pair and updates the editable matrix $E$ for CoCoBMs according to Equation~\ref{eq:editable_matrix}. We prompt an LLM with a concept $c_k$ and a label $y_j$ using a multiple-choice question (MCQ) to assess the relevance of $c_k$ to images annotated with $y_j$. The response options are: \textit{critical feature} ($c_k \in y_j$), \textit{occasionally present} ($c_k \in y_j$), and \textit{unrelated} ($c_k \notin y_j$). The matrix entry $E_{jk}$ is set to 1 if the concept is judged as either a critical feature or occasionally present, and 0 otherwise.

\noindent \textbf{Instance Selection.} To build a few-shot environment that enhances perceptual efficiency and better reflects real-world scenarios, we extract representative samples from the training set. These instances are selected via K-Means Clustering \citep{ilprints778} applied to the image features $\mathcal{I}(\mathcal{X})$, yielding $\beta$ clusters per label. The cluster centroids are used as $\beta N$ fixed instances across iterations, thereby stabilizing the grounding process.

\noindent \textbf{Environment Perception.} It employs the proposed CoCoBMs as a tool to interact with the environment, represented by the pre-selected instances. These instances are used to optimize the parameters of CoCoBMs, with the resulting validation set scores serving as environmental feedback. Notably, this feedback depends solely on the concept scores and is independent of the image labels.

\subsection{Feedback-based Concept Bank Planning}

The agent evaluates concept scores on the validation set as feedback, treating each concept as an atomic unit to assess its contribution within the overall concept bank. This analysis allows the planning module to identify and remove redundant concepts, while detecting gaps where certain labels lack identifiable concepts. These insights guide the action module in iteratively refining the concept bank to address such deficiencies.

\noindent \textbf{Score Activation Pattern.} For each concept $c$, let $S_{c} = \{s_i^j\} \in \mathcal{R}^{K \times N}$ denote its contribution scores on the validation set, where $s_i^j$ is the score of the $i$th sample for the $j$th label, $K$ is the number of validation samples, and $N$ is the number of labels. We first normalize the scores of each sample to the range $[-1, 1]$ using Equation~\ref{eq:score_norm}, categorizing the concept’s contribution to each label as positive, negative, or neutral.
\begin{equation}
    \label{eq:score_norm}
    \tilde{s}_i^j =
    \begin{cases} 
    \frac{s_i^j}{\max\{s_i^n \bigm| s_i^n > 0,\; n \in [1, N]\}}, & \text{if } s_i^j > 0 \\
    \frac{s_i^j}{\max\{|s_i^n| \bigm| s_i^n < 0,\; n \in [1, N]\}}, & \text{if } s_i^j < 0 \\
    0, & \text{if } s_i^j = 0
    \end{cases}
\end{equation}
We then compute the average normalized score for each label to obtain the score pattern $\mathcal{P}_{sc}^{c}$:
\begin{equation}
    \label{eq:score_mean}
    \mathcal{P}_{sc}^{c} = \left[\bar{s}_c^1, \ldots, \bar{s}_c^N \right], \text{where } \bar{s}_c^j = \frac{1}{K} \sum_{i=1}^{K} \tilde{s}_i^j
\end{equation}
The final binary score activation pattern $P_{act}^{c} = [a_1^{c}, \ldots, a_N^{c}]$ is obtained by thresholding:
\begin{equation}
    \label{eq:score_activation}
    a_j^{c} =
    \begin{cases} 
    1, & \text{if } \bar{s}_c^j > t_a \\ 
    0, & \text{otherwise}
    \end{cases}
\end{equation}
where $t_a \in [0, 1]$ is a threshold controlling concept activation. When $t_a = 1$, only the label with the highest contribution is activated; when $t_a = 0$, all labels with non-zero contributions are retained. This label-specific activation is unique to CoCoBMs, as original CBMs share concept scores across all labels, making it impossible to isolate contributions at the label level.

\noindent \textbf{Redundant Concept.} We categorize redundancy into two cases: (1) the concept does not contribute to any label when activated (i.e., $\sum_{j=1}^{N} a_j^{c}=0$); (2) for a concept $c_i$, there exists another concept $c_j$ with an identical binary activation pattern ($P_{act}^i = P_{act}^j$). To assess redundancy, we compute the Hadamard product of each concept’s activation pattern and contribution scores (i.e., $P_{sc}^i \cdot P_{act}^i$ and $P_{sc}^j \cdot P_{act}^j$) and calculate the Manhattan distance between them. If the distance is below a threshold $t_m$ and $c_i$ has a lower total positive contribution, defined as the sum of elements in the Hadamard product (i.e., $\sum(P_{sc}^i \cdot P_{act}^i) < \sum(P_{sc}^j \cdot P_{act}^j)$), then $c_i$ is considered redundant and only $c_j$ is retained.

\noindent \textbf{Insufficient Concept.} For each label, we analyze its support from the current concept set. A label is deemed unidentifiable if: (1) no concept is activated for it; or (2) it shares identical set of activated concepts with another label. These labels are forwarded to the action module to guide the generation of additional concepts.

If no missing concepts are detected, the agent terminates the iteration process and trains CoCoBMs on the full dataset to obtain the final performance.

\section{Experiments}

\begin{figure*}[t]
  \centering
  \includegraphics[width=0.8\linewidth]{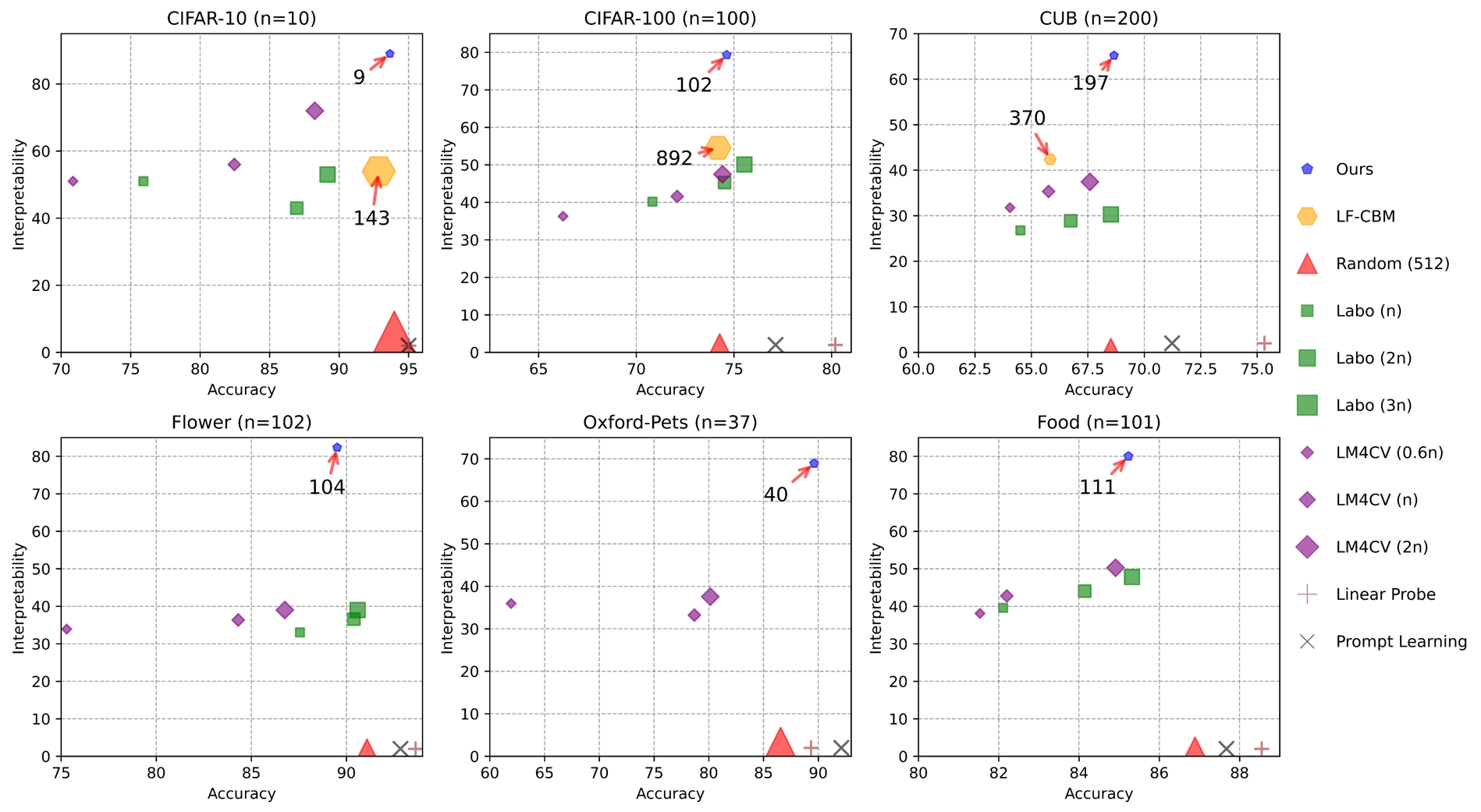}
  \caption{Comparison with state-of-the-art CBMs and black-box models. The legend follows the format: method (\#concepts used). For example, \textit{LM4CV (2n)} indicates using twice \#labels as concepts for each dataset. Red arrows mark the \#concepts used by our method and \textit{LF-CBM}. Marker size reflects the relative size of each concept bank.}
  \label{fig:main_exp}
\end{figure*}

\subsection{Datasets}
We evaluate and benchmark our approach on 6 datasets of diverse scales and challenges, following the dataset partitioning strategy of \citet{yan2023learning}: CUB \citep{wah2011caltech}, CIFAR-10 and CIFAR-100 \citep{krizhevsky2009learning}, Food-101 \citep{bossard2014food}, Flower \citep{nilsback2008automated} and Oxford-Pets \citep{parkhi2012cats}.

\subsection{Evaluations}
Performance is evaluated in terms of accuracy and interpretability. Accuracy is measured using the standard classification metric, and interpretability is quantitatively assessed via a novel LLM-based approach.

\noindent \textbf{Interpretability.} We define positively contributing concept scores as reasoning evidence that supports the model’s prediction, offering interpretable insights for humans. Interpretability is evaluated at the label level from two complementary aspects: \textit{truthfulness} and \textit{distinguishability}.

Let $\hat{y_j}=\{s_{c_k}^j \}$ denote the predicted label $\hat{y_j}$ of a given sample along with the corresponding concept scores, where $s_{c_k}^j$ is the score of concept $c_k$ for the predicted $j$th label, and $k \in \{1, \dots, M\}$. For scores where $s_{c_k}^j > 0$, we apply local min-max normalization and set all non-positive scores to zero. Next, we compute the mean of the normalized scores across all validation samples to obtain a global contribution profile for each label. Global min-max normalization is then applied to the aggregated concept scores, and concepts are ranked by their normalized contributions $\tilde{s}_{c_k}^j$. The final explanation for label prediction $\hat{y}_j$ across the dataset is thus represented as an ordered list of concepts $[c_1, \dots, c_p]$, where $p \leq M$.

\noindent \textbf{Truthfulness.} This metric evaluates whether the concepts that support the predicted labels are consistent with objective real-world facts. To better reflect practical reasoning, the evaluation focuses on combinations of relevant concepts rather than individual ones. Given that concepts vary in importance, we define a set of thresholds $t_c = \{0, 0.25, 0.50, 0.75, 1\}$ to enable hierarchical evaluation based on contribution strength. At each threshold level, we select the subset of concepts $\left[c_1, \dots, c_p\right]$ satisfying $\tilde{s}_{c_k}^j > t_c$. When $t_c = 1$, only the most impactful concepts are evaluated; when $t_c = 0$, all positively contributing concepts are included. For each threshold, we construct an MCQ to prompt an LLM for judgment. Each MCQ provides two options: (1) the concept combination aligns with objective facts; or (2) most concepts are irrelevant or contradictory.

\noindent \textbf{Distinguishability.} This metric assesses whether the provided concepts can effectively distinguish among labels. We prompt an LLM with a concept set and a list of labels to identify the most appropriate one. To construct distractor options, we first compute textual similarity between label names using RoBERTa embeddings \citep{liu2019roberta}, selecting the top 8 most similar labels. Likewise, visual similarity is computed using the CLIP image encoder by averaging image representations per label. Based on these two similarity rankings, we create two MCQs per modality: one using the top 4 and another using the bottom 4 similar labels as distractors, with option order randomized. An additional MCQ includes 4 randomly sampled labels and the correct answer. In total, each evaluation consists of 5 MCQs, each with 5 options.

Thus, interpretability for each label is assessed using 10 MCQs, with 5 for truthfulness and 5 for distinguishability. The final score is calculated as the arithmetic mean of these two metrics. To ensure fairness and reproducibility, all MCQs remain fixed for each dataset during evaluation.

\subsection{Implementation Details}
We use CLIP ViT-B/32 as the backbone for CoCoBMs and all baseline models. Following \citet{zhou2022learning}, the number of learnable tokens in conditional learning is set to 8. The batch size is configured to 2,048 across all datasets. All models are trained using the Adam optimizer \citep{diederik2014adam} with a constant learning rate of 0.01.

The Concept Agent prompts the GPT-4o API \citep{hurst2024gpt} for concept generation and verification. The number of selected concepts equals the number of prompt labels. In the few-shot feedback phase, 16 samples are used as instances. A threshold of $t_a = 0.1$ is empirically chosen to identify feedback-activated concepts. Concept pairs with a Manhattan distance below $t_m = 0.3$ are considered redundant. For evaluation, GPT-4-turbo is prompted to answer MCQs, each repeated 3 times, with the majority vote taken as the final result.

\subsection{Baselines} We compare our approach with SOTA CBMs that construct concept banks using LLMs, including Label-free CBMs \citep{oikarinen2023label}, LaBo \citep{yang2023language} and LM4CV \citep{yan2023learning}. To illustrate that CBMs can achieve high accuracy without interpretability given a sufficiently large concept set, we also include random-word concept banks as a non-interpretable baseline. For LaBo, we experiment with 1, 2 and 3 concepts per label. For concise LM4CV, we build concept banks sized at 0.6×, 1×, and 2× the number of labels. Publicly released concept banks provided by these work are used. As black-box baselines, we include image feature-based linear probes and CLIP-based prompt learning \citep{zhou2022learning} with 8 learnable tokens.

\begin{figure}[!t]
  \centering
  \includegraphics[width=0.95\columnwidth]{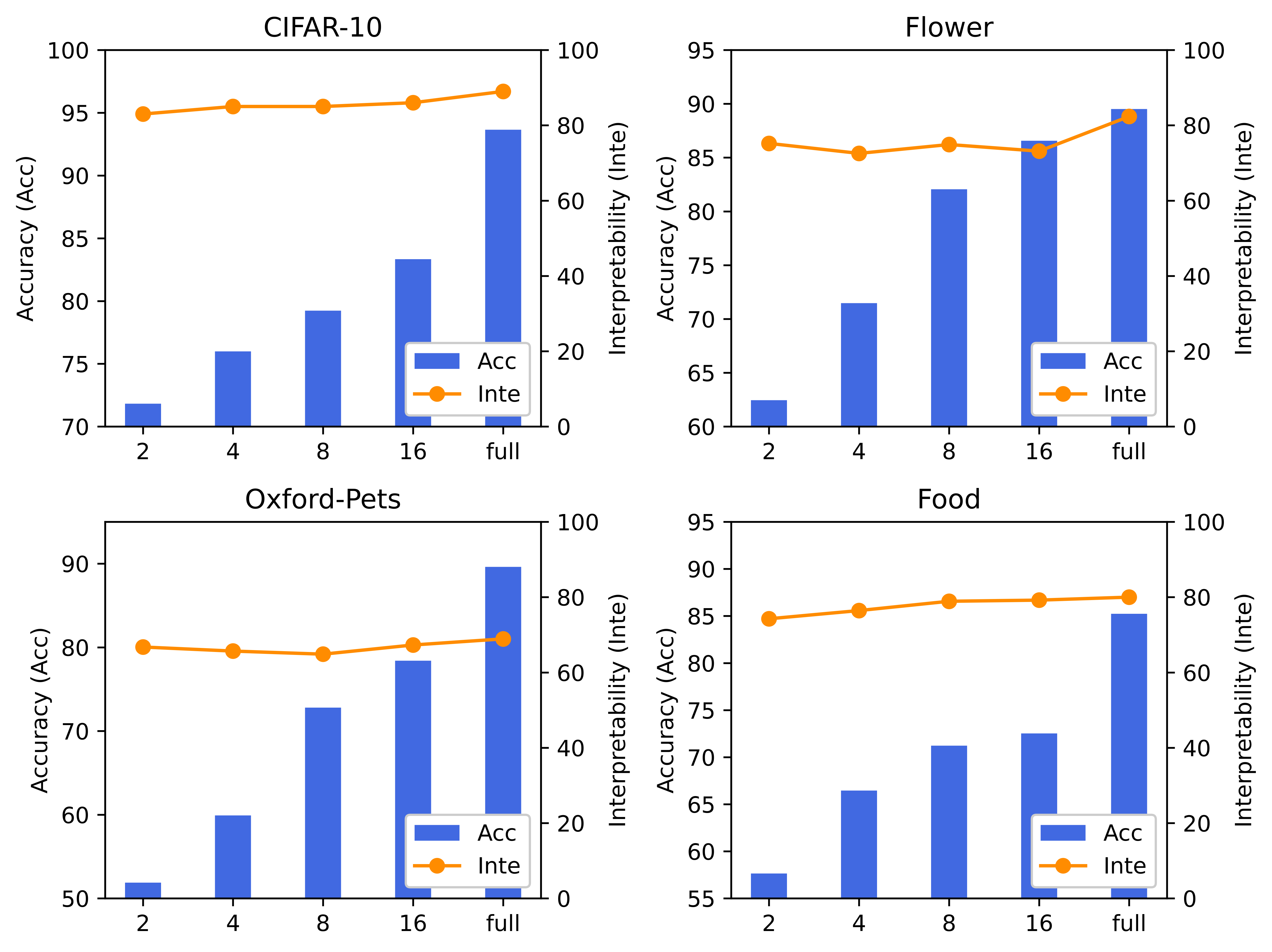}
  \caption{Effect of sample size on the classification accuracy (Acc\%) and interpretability (Inte\%) of CoCoBMs. The x-axis indicates the number of samples.}
  \label{fig:few_shot}
\end{figure}

\subsection{Accuracy vs. Interpretability Trade-Off}
Figure~\ref{fig:main_exp} shows the evaluation results of our Concept Agent in comparison with CBM baselines and black-box models across six datasets.

\noindent \textbf{Accuracy.} Under our configuration, the number of concepts determined by the Concept Agent approximately equals the number of labels. For a fair comparison, we evaluate against LaBo-$n$ and LM4CV-$n$, where $n$ denotes the number of labels. Our method achieves an average accuracy gain of 6.15\% over LaBo-$n$ across five datasets. Remarkably, it still outperforms LaBo-3$n$ by 0.51\% on average, despite LaBo using three times as many concepts. Similarly, our approach outperforms LM4CV-$n$ by 5.97\% and LM4CV-2$n$ by 3.21\%. Compared to LF-CBM, which uses significantly more concepts (e.g., 16× on CIFAR-10), our method achieves a 1.36\% higher average accuracy. These results demonstrate that our method delivers superior classification performance with a much more compact concept bank. Our method narrows the performance gap between CBM-style models and black-box models, reducing it to 3.45\% relative to linear probing and 2.44\% relative to prompt learning.

\noindent \textbf{Interpretability.} Our approach also substantially enhances interpretability, achieving an average score of 77.46\%, with truthfulness and distinguishability scores of 81.59\% and 73.34\%, respectively. This represents an approximately 30\% improvement over LM4CV-2$n$, demonstrating superiority over existing CBM-based models. We also show that CBMs can still attain strong classification accuracy when using a concept bank composed of random words with 512 concepts, emphasizing the need for rigorous interpretability evaluation.

It reveals that our method consistently trends towards the top-right region across all datasets, reflecting a better trade-off between accuracy and interpretability than existing SOTA CBMs.

\begin{figure*}[t]
  \centering
  \includegraphics[width=1\linewidth]{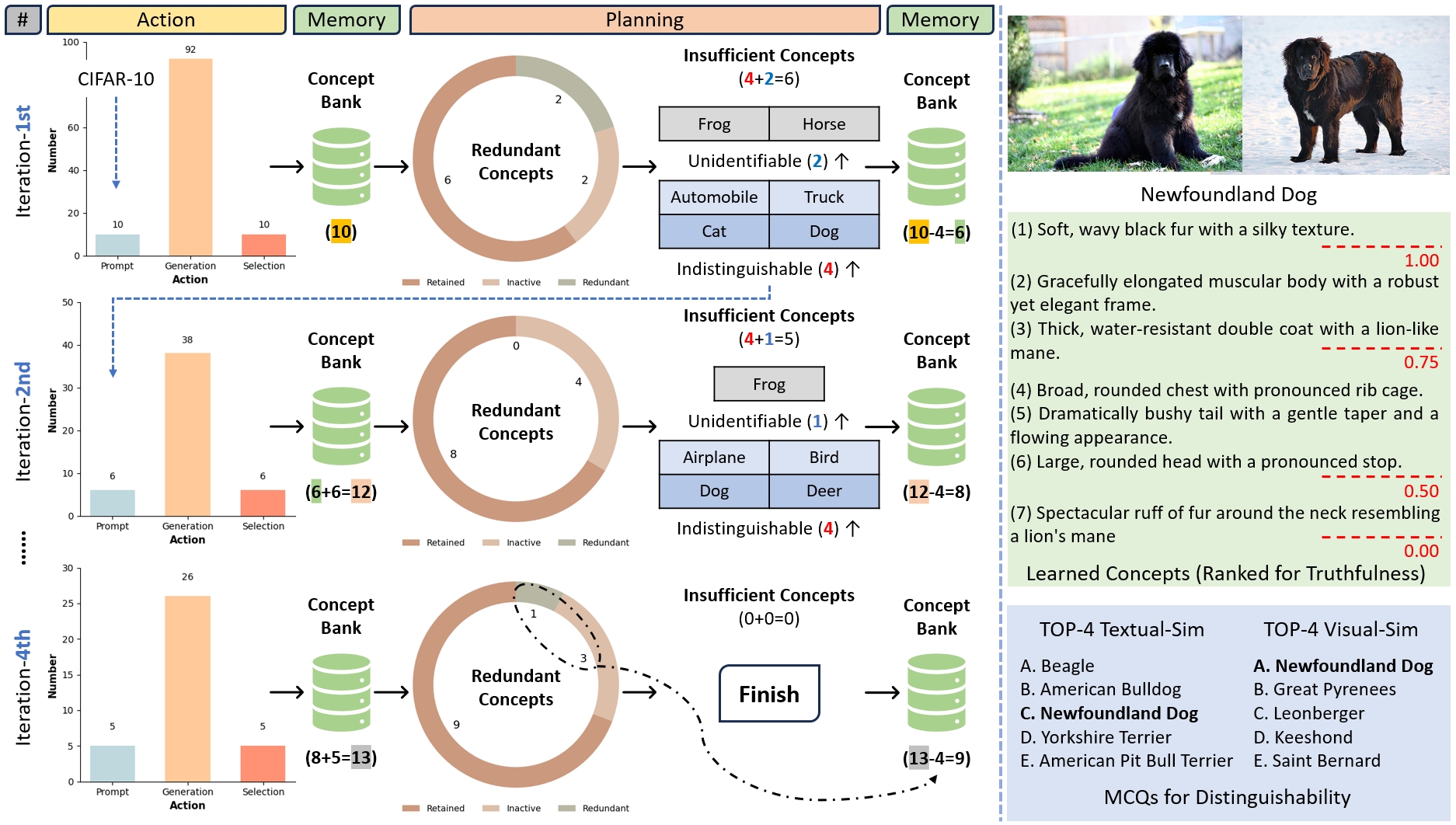}
  \caption{\textbf{Case Study.} \textbf{\textit{Left}}: Concept bank evolution on CIFAR-10 (3rd iteration omitted), concluding after 4 rounds with 9 final concepts. The x-axis of the Action module denotes the number of labels used to \textit{prompt} the LLM, total \textit{generated} concepts, and \textit{selected} concepts. \textbf{\textit{Right}}: Label-level concepts for \textit{Newfoundland Dog} in Oxford-Pets, ranked by normalized contribution. \textbf{\textit{Bottom right}}: MCQs with distractors based on textual and visual similarity.}
  \label{fig:case_study}
\end{figure*}

\subsection{Ablation Study}

\noindent \textbf{Interpretability in Few-shot Learning.} The agent refines the concept bank through feedback-driven optimization in a few-shot environment. As shown in Figure~\ref{fig:few_shot}, evaluation on few-shot samples using the finalized concept bank demonstrates that accuracy improves with increasing sample size, while interpretability remains stable with only minor fluctuations. These results highlight the model's robustness in maintaining interpretability under limited data conditions, while enhancing perceptual efficiency during feedback.

\begin{table}
  \centering
  \resizebox{0.95\columnwidth}{!}{%
  \begin{tabular}{ccc}
   \hline
    \textbf{Dataset} & \textbf{Acc (Sta → Dyn)} & \textbf{Inte (Sta → Dyn)} \\
    \hline
    CIFAR-100 & 72.67 $\rightarrow$ 74.63 & 67.90 $\rightarrow$ 79.30 \\
    Flower & 87.45 $\rightarrow$ 89.51 & 70.59 $\rightarrow$ 82.35 \\
    Food & 85.31 $\rightarrow$ 85.23 & 70.89 $\rightarrow$ 80.00 \\
    \hline
  \end{tabular}%
  }
  \caption{Accuracy (Acc\%) and interpretability (Inte\%) comparison between static (Sta) grounding and dynamic (Dyn) grounding. Static grounding is based on the initialized concept bank in a one-directional workflow.}
  \label{tab:static_comp}
\end{table}

\begin{table}
  \centering
  \resizebox{\columnwidth}{!}{%
  \begin{tabular}{ccc}
   \hline
    \textbf{Dataset} & \textbf{Acc (w/ E → w/o E)} & \textbf{Inte (w/ E → w/o E)} \\
    \hline
    CIFAR-100 & 74.63 $\rightarrow$ 76.95 & 79.30 $\rightarrow$ 39.60 \\
    Flower & 89.51 $\rightarrow$ 89.61 & 82.35 $\rightarrow$ 35.59 \\
    Oxford-Pets & 89.62 $\rightarrow$ 89.94 & 68.92 $\rightarrow$ 39.46 \\
    \hline
  \end{tabular}%
  }
  \caption{Ablation results on accuracy (Acc\%) and interpretability (Inte\%) between CoCoBMs with and without editable matrix ($E$) across three datasets.}
  \label{tab:editable_comp}
\end{table}

\noindent \textbf{Dynamic Grounding vs. Static Grounding.} To assess the effectiveness of dynamic grounding, we compare the adaptively refined concept bank with its initial static version. As shown in Table~\ref{tab:static_comp}, incorporating environment feedback significantly enhances interpretability, yielding an average improvement of 10.76\%. While a slight drop in accuracy is observed on the Food dataset, the overall classification accuracy improves across datasets.

\noindent \textbf{Editable Matrix.} We evaluate the effect of removing the editable matrix from CoCoBMs. As shown in Table~\ref{tab:editable_comp}, the editable matrix slightly constrains accuracy but substantially improves interpretability by incorporating factual knowledge from LLMs. Without the editable matrix, the interpretability of our method becomes comparable to baseline models. These results suggest that while condition learning and category-specific scoring enhance classification performance, they contribute less to interpretability in the absence of factual constraints.

\subsection{Case Study}

Figure~\ref{fig:case_study} illustrates the iterative process by which the Concept Agent dynamically grounds a concept bank on the CIFAR-10 dataset. After four iterations, the final bank contains nine concepts. During this process, inactive concepts, those with minimal contributions to any label, are removed. This reveals a limitation of LLM-based static concept grounding: although concepts are derived from label names, some do not appear in downstream images due to dataset bias or remain undetected due to CLIP pretraining bias. We also observe that some redundant concepts are removed due to identical activation patterns and mutually low Manhattan distances, suggesting that existing selection algorithms may still yield functionally similar concepts.

For categories with insufficient concepts, we identify semantically similar label pairs that are difficult to distinguish and align with human perception, such as \textit{cat versus dog} (both pets), \textit{automobile versus truck} (both vehicles), and \textit{bird versus airplane} (visually similar). These findings demonstrate the agent’s ability to uncover confounding label pairs that can guide further refinement and expansion of the concept bank. We additionally observe that removing inactive concepts can leave some labels without any associated concepts. The agent detects and compensates for these missing-label cases, highlighting its capacity to repair and maintain a complete concept bank.

The right side of Figure~\ref{fig:case_study} presents a case study of the \textit{Newfoundland Dog} from the Oxford-Pets dataset, illustrating seven concepts that are both factually grounded and contribute to its recognition. Concepts are ranked by normalized contribution scores, and thresholds are defined to enable hierarchical evaluation of truthfulness. The bottom-right section presents distractors of varying difficulty, generated via text and image similarity, posing greater challenges to distinguishability evaluation.

\section{Conclusion}
We present an LLM-driven Concept Agent that dynamically adjusts the concept bank based on environmental feedback to determine the optimal number of concepts. The agent uses our proposed CoCoBMs as a tool to perceive, enabling concept-based interpretable image classification. Our approach not only improves classification accuracy but also significantly enhances the interpretability through a novel quantitative evaluation metric.

\section*{Limitations}
Our approach utilizes open-source LLMs for concept generation. However, evaluating the internal knowledge of LLMs and managing the inherent randomness in concept generation, both of which may affect the performance and evaluation of concept agents, remains open challenges.

In the fact verification phase, all possible concept–category pairs are validated, which limits the scalability of our method. This limitation stems from the nature of traditional CBMs, which require scoring all concepts in the bank to produce final predictions. To mitigate this, we explored a filtering strategy using CLIP’s modality alignment to preselect concept–category pairs for verification. However, our experiments showed that this approach substantially increases the number of agent iterations, leading to higher computational costs compared to exhaustive enumeration.

\bibliography{custom}

\clearpage

\appendix

\onecolumn

\section{Prompt Templates}
\label{sec:appendix_prompts}

\definecolor{CoolAccent}{HTML}{FF5575}
\begin{tcolorbox}[title=Prompt for Concept Generation (Initialization and Unidentifiable Labels.), left=2mm,right=1mm,top=0mm, bottom=0mm,colback=white,colframe=CoolAccent]
    \begin{lstlisting}[style=plain]
What are the helpful visual features to distinguish "[class name]" from other "[superclass]"?

Each feature should be a longish modifier noun phrase. The noun should represent a single visually observable aspect, depicting one characteristic or attribute. The modifiers should be rich and specific, highlighting the unique presentation of this aspect and avoiding vague terms like "distinctive" or "signature".

Do not use the word "[class name]" or any specific instance names from "[class name]". 

List each feature on a new line with no additional content or numbering.

Note: Please ensure that your listed features do not overlap with the following features:
    \end{lstlisting}
    \end{tcolorbox}
\begin{tcolorbox}[title=Prompt for Concept Generation (Indistinguishable Labels), left=2mm,right=1mm,top=0mm, bottom=0mm,colback=white,colframe=CoolAccent]
    \begin{lstlisting}[style=plain]
What are the helpful visual features to distinguish between "[class name list]"?

Each feature should be a longish modifier noun phrase. The noun should represent a single visually observable aspect, depicting one characteristic or attribute. The modifiers should be rich and specific, highlighting the unique presentation of this aspect and avoiding vague terms like "distinctive" or "signature".

List each feature on a new line with no additional content or numbering.

Note: Please ensure that your listed features do not overlap with the following features:
    \end{lstlisting}
    \end{tcolorbox}

\definecolor{CoolAccent}{HTML}{14BC94}
\begin{tcolorbox}[title=Prompt for Fact Verification, left=2mm,right=1mm,top=0mm, bottom=0mm,colback=white,colframe=CoolAccent]
    \begin{lstlisting}[style=plain]
Is the phrase "[concept]" a feature that helps identify the presence of "[class name]" in photos?

Select the most appropriate option without providing an explanation.

A. This feature is critical and highly prominent.
B. This feature may occasionally appear, but it is typically not significant.
C. This feature is unrelated to the described object and unhelpful for identification.
    \end{lstlisting}
    \end{tcolorbox}

\definecolor{CoolAccent}{HTML}{6299FF}
\begin{tcolorbox}[title=Prompt for Evaluation (Truthfulness)
, left=2mm,right=1mm,top=0mm, bottom=0mm,colback=white,colframe=CoolAccent]
    \begin{lstlisting}[style=plain]
I have a batch of images of "[class name]". Someone has summarized several critical features, ranked by prominence (with the most prominent features listed first) for recognizing "[class name]":
"[feature list]"

Please evaluate whether the summarized features align with objective facts or real-world knowledge? Select the most appropriate option without providing an explanation.

A. Overall aligns with facts.
B. Most features do not align with facts or are contradictory to each other.
    \end{lstlisting}
    \end{tcolorbox}

\begin{tcolorbox}[title=Prompt for Evaluation (Distinguishability),
 left=2mm,right=1mm,top=0mm, bottom=0mm,colback=white,colframe=CoolAccent]
    \begin{lstlisting}[style=plain]
I have a batch of images characterized by the following features, ranked by prominence (with the most prominent features listed first):
"[feature list]"

Which of the following "[superclass]" is most likely to appear in these images? Please select the most appropriate answer without providing an explanation.

A. [A]; B. [B]; C. [C]; D. [D]; E. [E]
    \end{lstlisting}
    \end{tcolorbox}

\vspace{0.5em}
\begin{minipage}{0.96\textwidth}
\captionsetup{type=figure}
\captionof{figure}{Prompt templates used in the Concept Agent’s action module, and interpretability evaluation templates.}
\end{minipage}

\end{document}